\documentclass[10pt,twocolumn,letterpaper]{article}

\usepackage{iccv}
\usepackage{times}
\usepackage{epsfig}
\usepackage{graphicx}
\usepackage{amsmath}
\usepackage{amssymb}
\usepackage{subcaption} % for 'subtable' env
\usepackage{multirow} 
\usepackage{bigstrut}
\usepackage{bm}
\usepackage{amsthm,amsmath,amssymb}
\usepackage{colortbl}  %彩色表格需要加载的宏包
\usepackage{xcolor}
\usepackage{array}
\usepackage{mathrsfs}
\usepackage{booktabs}

\usepackage[pagebackref=true,breaklinks=true,letterpaper=true,colorlinks,bookmarks=false]{hyperref}

\iccvfinalcopy % *** Uncomment this line for the final submission

\def\iccvPaperID{5674} % *** Enter the ICCV Paper ID here

\ificcvfinal\fi

\begin{document}

%%%%%%%%% TITLE
\title{Pixel Adaptive Deep Unfolding Transformer for \\Hyperspectral Image Reconstruction}

\author {
    % Authors
    Miaoyu Li\textsuperscript{\rm 1},  
   Ying Fu\textsuperscript{\rm 1}\thanks{ Corresponding Author}, 
   Ji Liu\textsuperscript{\rm 2}, 
    Yulun Zhang\textsuperscript{\rm 3}\\
    \textsuperscript{\rm 1}Beijing Institute of Technology, \textsuperscript{\rm 2}Baidu Inc., \textsuperscript{\rm 3}ETH Z\"{u}rich \\
    {\tt\small miaoyli@bit.edu.cn,  fuying@bit.edu.cn, liuji04@baidu.com, yulun100@gmail.com}\\ 
}

\maketitle
\ificcvfinal\thispagestyle{empty}\fi

%%%%%%%%% ABSTRACT
\begin{abstract}
   Hyperspectral Image (HSI) reconstruction has made gratifying progress with the deep unfolding framework by formulating the problem into a data module and a prior module. Nevertheless, existing methods still face the problem of insufficient matching with HSI data. The issues lie in three aspects: 1) fixed gradient descent step in the data module while the degradation of HSI is agnostic in the pixel-level. 2) inadequate prior module for 3D HSI cube. 3) stage interaction ignoring the differences in features at different stages. To address these issues, in this work, we propose a Pixel Adaptive Deep Unfolding Transformer (PADUT) for HSI reconstruction. In the data module, a pixel adaptive descent step is employed to focus on pixel-level agnostic degradation. In the prior module, we introduce the Non-local Spectral Transformer (NST) to emphasize the 3D characteristics of HSI for recovering. Moreover, inspired by the diverse expression of features in different stages and depths, the stage interaction is improved by the Fast Fourier Transform (FFT). Experimental results on both simulated and real scenes exhibit the superior performance of our method compared to state-of-the-art HSI reconstruction methods. The code is released at: \textcolor{magenta}{https://github.com/MyuLi/PADUT}
\end{abstract}

%%%%%%%%% BODY TEXT
\section{Introduction}

Hyperspectral Images (HSIs) have been widely
applied in multiple fields, \eg, material identification~\cite{keshava2004distance,khan2018modern}, spectral unmixing~\cite{jia2008constrained}, and medical analysis~\cite{zhi2007classification,fei2019hyperspectral}. Conventional hyperspectral
imagers are time-consuming and inflexible when scanning a scene to capture the target HSI.
To overcome this limitation and capture the diverse reluctance of a real scene, coded aperture snapshot spectral imaging (CASSI) modulates 3D HSI into a compressed 2D measurement by a coded aperture and a disperser~\cite{wagadarikar2008single}. Since the image quality of desired 3D HSI is restricted by the performance of reconstruction algorithms, exploring effective reconstruction algorithms is of much importance.

%Since the imaging quality is hugely , reconstruction algorithm is important for the CASSI system. Therefore, 

% However, real HSIs often suffer from different kinds of
% degradations, \ie, noise, undersampling, data missing,  especially under non-ideal imaging conditions~\cite{manea2015hyperspectral} or data transmission procedures~\cite{schweizer2001efficient,dua2020comprehensive}. Thus, HSI reconstruction is a fundamental step, which not only improves the visual quality of HSIs but also  facilitates the subsequent applications.

\begin{figure}[t]
\scriptsize 
\centering	
\includegraphics[width=1\linewidth]{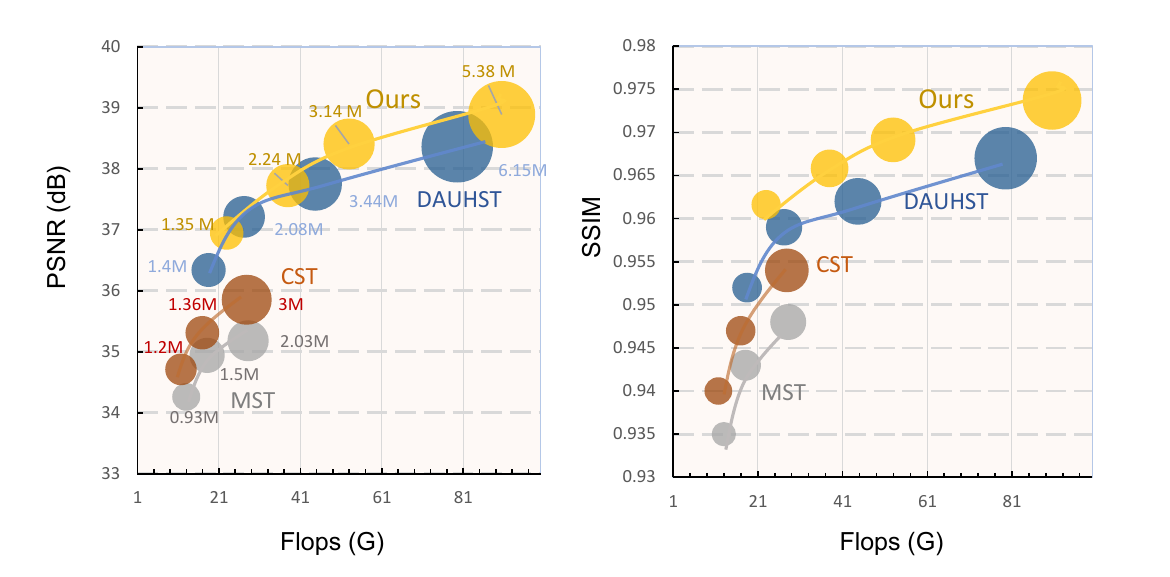}
\vspace{-3mm}                                  
\caption{Performance vs. GFLOPs. Under different model sizes, PADUT outperforms the state-of-the-art methods.}
\vspace{-2mm}
\label{fig:performance}
\end{figure}
To address the ill-posed reverse problem of HSI reconstruction, traditional model-based methods explore various priors in different solution space with interpretability. The widely employed priors can be summarized as non-local similarity~\cite{fu2016exploiting,zhang2019computational}, low-rank property~\cite{liu2018rank}, sparsity~\cite{zhang2018spectral} and total variation~\cite{eason2014total,yuan2016generalized}, \emph{etc}. However, these hand-crafted priors have limited generalization ability and often result in a mismatch between prior assumptions and the problem. Consequently, such methods cannot characterize the features of HSI under various scenarios and typically require time-consuming numerical iteration.

Recently, deep convolutional neural networks~\cite{yuan2018hyperspectral,miao2019net,cai2022mask,meng2020end}  
have been applied for HSI reconstruction and achieved decent performance. According to different learning strategies, deep learning-based methods can be broadly categorized into two groups, \ie, end-to-end learning methods~\cite{yuan2018hyperspectral,meng2020end} and model-aided methods~\cite{xiong2021mac,cai2022degradation}. End-to-end learning methods recover the original HSI via brute-force mapping to learn the spatial and spectral information. Some of these typical works include $\lambda$-Net~\cite{miao2019net}, TSA-Net~\cite{meng2020end}, and MST~\cite{cai2022mask}. Without the guidance
of physical models, these end-to-end learning methods are black boxes and lack transparency.
In contrast, the model-aided methods~\cite{xiong2021mac} leverage the physical characteristics of HSI degradation into deep networks, resulting in inherent interpretability. The most well-known model-aided approaches include Plug-and-Play (PnP) \cite{meng2021self} and Deep Unfolding \cite{huang2021deep,wang2020dnu,cai2022degradation}. Since PnP-based methods generally exploit a fixed pretrained denoiser and fail to learn a specific mapping for HSI, their reconstruction performance is hampered. Recently, deep unfolding-based methods~\cite{wang2020dnu,huang2021deep,cai2022degradation} have achieved significant improvements in HSI reconstruction. 

\begin{figure}[t]
\scriptsize 
\centering	
\includegraphics[width=0.96\linewidth]{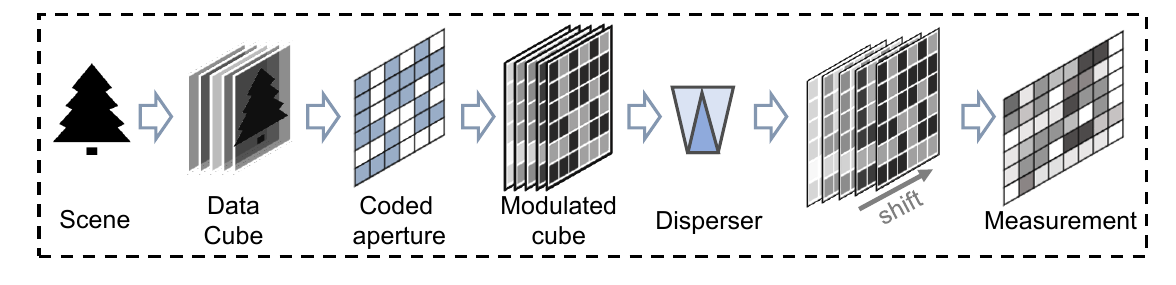}
\vspace{-2mm}
\caption{The coded aperture snapshot spectral imaging (CASSI) system. Different positions in the coded measurement may suffer from different levels of degradation. }
\label{fig:cassi}
\end{figure}

\begin{figure}[]
\scriptsize 
\centering	
\includegraphics[width=0.95\linewidth]{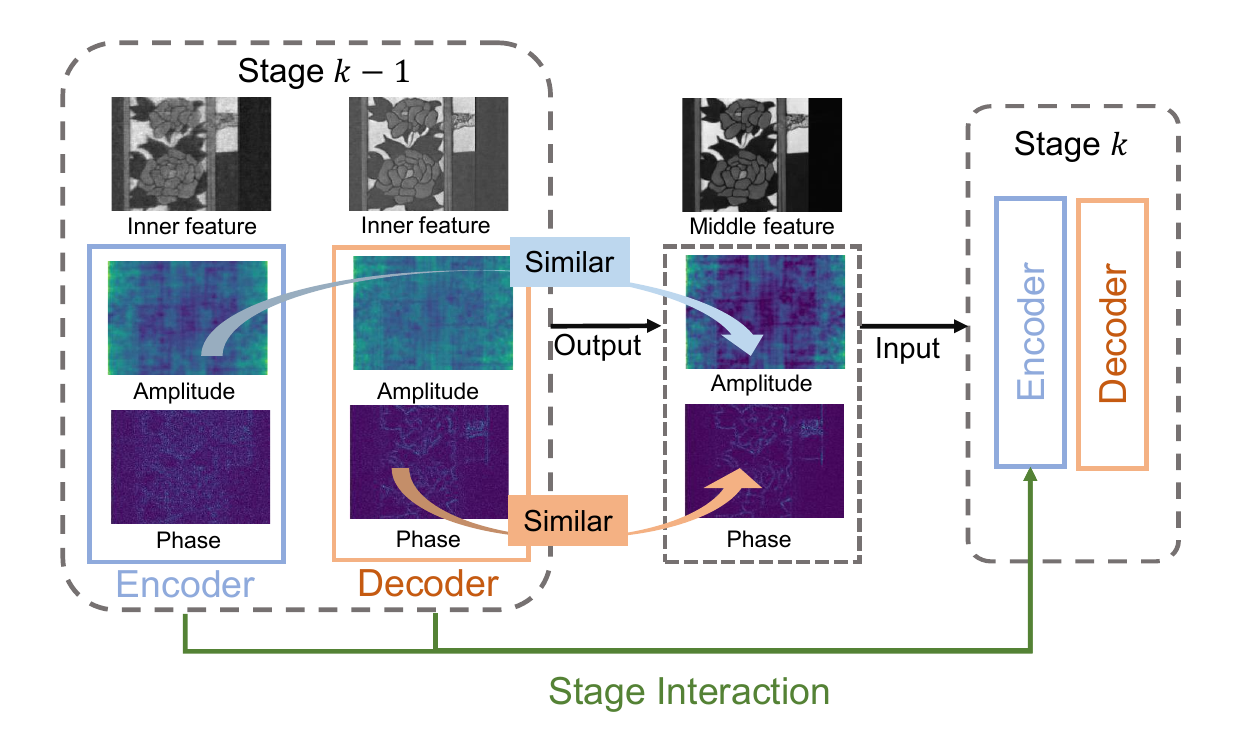}
\vspace{-2mm}
\caption{Visualization of inner features of deep unfolding networks in the frequency domain. The Amplitude/Phase image is extracted by the inner feature with Phase/Amplitude component set to a constant.}
\vspace{-2mm}
\label{fig:fft}
\end{figure}

As deep unfolding framework unfolds the iterative optimization algorithms with neural networks, it typically includes a data module and a prior module. As illustrated in Figure \ref{fig:cassi}, pixels in the 3D cube at different positions are compressed agnostically in the measurement, while the existing algorithm ignores such pixel-specific degradation in the data module. For the prior module, denoiser plays a critical role in the multi-stage optimization. Since HSIs are in 3D representation, it remains an issue for existing denoisiers to effective use of spatial-spectral information. 

Besides, during the iterative recovery process, cross-stage fusion is necessary to prevent the loss of key information while gaining comprehensive features. We observe that the frequency information varies at different stages and depths (see Figure \ref{fig:fft}). The features from the previous encoder layer have clearer amplitude information, while the features from the previous encoder layer have clearer phase information. However, in the former works~\cite{wang2020dnu,zhang2022herosnet,cai2022degradation}, the underlining differences of cross-stage features in the frequency domain are ignored, which incurs inferior performance of the multi-stage framework.

Inspired by the above findings, we propose a \textbf{P}ixel \textbf{A}daptive \textbf{D}eep \textbf{U}nfolding \textbf{T}ransformer (PADUT) framework for HSI reconstruction. \textbf{First}, we introduce the deep unfolding framework for the reconstruction process with the position-specific degradation information taken into consideration in the data module. \textbf{Second}, we propose a Non-local Spectral reconstruction Transformer for HSI to utilize the two-dimensional data of HSI in each stage.  \textbf{Third}, we employ the frequency component analysis of HSI to fuse the features across iterative stages. With the observation that the encoder feature and the decoder feature have different emphases in the frequency domain, we propose Fast Fourier Transform Stage Fusion (FFT-SF) module, which leads to more comprehensive features for superb performance. The specific contributions of our work are:
\vspace{-1.5mm}
\begin{itemize}
\setlength{\parsep}{0pt}
\setlength{\parskip}{0pt}
\item We propose a pixel adaptive deep unfolding transformer for HSI reconstruction. In the data module, we introduce pixel-level adaptive recovery at different locations. In the prior module, we propose a Non-local Spectral transformer for HSI processing.
\vspace{1mm}
\item We introduce a novel frequency perspective for the cross-stage features in the iterative reconstruction process. Particularly, amplitude and phase representations are employed to establish interactions between different stages and depths.
\vspace{1mm}
\item We carry out extensive experiments on both simulation scenes and real scenes to exhibit the effectiveness of our method for HSI reconstruction.
\end{itemize}

\section{Related Work}
In this section, we provide a brief overview of the recent advances in the field of HSI construction. Recent years have witnessed a paradigm shift from model-based methods to deep learning methods. Now, a progressive practice is to incorporate physical constraints with a data-driven network.
\subsection{Model-based methods}
Model-based HSI reconstruction methods~\cite{bioucas2007new,yuan2016generalized,wang2016adaptive,zhang2019computational,liu2018rank,zhang2018fast,he2020non} commonly optimize the objective function by separating the data fidelity term and regularization term. Twist~\cite{bioucas2007new} introduced a two-step Iterative shrinkage/thresholding algorithm for reconstruction
with missing samples. The non-local similarity and low-rank regularization were utilized in ~\cite{fu2016exploiting} and \cite{he2020non}. The sparse representation~\cite{lin2014spatial,wang2016adaptive} and Gaussian mixture model~\cite{yang2014compressive}  have also been widely studied. In \cite{yuan2016generalized}, generalized alternating projection (GAP) method was proposed for HSI compressive sensing with utilized the total variation minimization. Although these methods
produce satisfying results in the case of proper situation, they still face the problems of lacking generalization ability and computational efficiency.

\subsection{Deep Learning-based methods}
 Inspired by the remarkable success of deep neural networks, deep learning-based methods~\cite{wang2019hyperspectral,fu2021coded,cai2022mask,lai2022deep} have gained widespread utilization in the HSI reconstruction area. The pioneering methods SDA~\cite{mousavi2015deep} and Reconnet~\cite{kulkarni2016reconnet} demonstrated the effectiveness of deep networks compared to traditional methods. Later, TSA-Net~\cite{meng2020end} introduced the spatial-spectral self-attention to sequentially reconstruct HSI. An external and internal learning method was introduced in ~\cite{fu2021coded} for utilizing the image-specific prior. To explore the power of Transformer for compressive sensing, MST~\cite{cai2022mask} and CST~\cite{yun2022coarse} were proposed to capture the inner similarity of HSIs. Hu \etal \cite{hu2022hdnet} introduced the high-resolution dual-domain learning network (HDNet) to solve the spectral compressive imaging task. Though significant progress has been made, it is difficult for these brute-force methods to utilize the physical degradation characteristics.

\subsection{Interpretable networks methods}

 Model-guided interpretable networks for HSI denoising have been actively explored in~\cite{wang2020dnu,bodrito2021trainable,xiong2021mac,cai2022mask,zhang2022herosnet}. On the one hand, traditional handcrafted prior is employed in deep networks as a component~\cite{qiu2021effective}.
 %\cite{bodrito2021trainable} proposed a trainable sparse coding model for HSI denoising. 
On the other hand, conventional optimization algorithms can be represented by recurrent deep networks, allowing for the application of deep learning techniques to optimization problems. The plug-and-play
 methods ~\cite{yuan2020plug,zheng2021deep} plugged the deep denoisers into the optimization process. 
Deep unfolding HSI reconstruction methods~\cite{wang2020dnu} enjoyed the
power of deep learning and known degradation
process. For instance, GAP-Net unfolded the generalized alternating projection algorithm with a trained convolutional neural network. DAUHST~\cite{cai2022degradation} introduced a novel half-shuffle Transformer into the unfolding framework.
Different from those methods that have limitations in exploring the pixel-specific degradation information and cross-stage features, our method introduces the pixel-adaptive recovery data module and utilizes the frequency information to guide the cross-stage fusion process.
\section{Method}

\begin{figure*}[t]
\scriptsize 
\centering	
\includegraphics[width=0.95\linewidth]{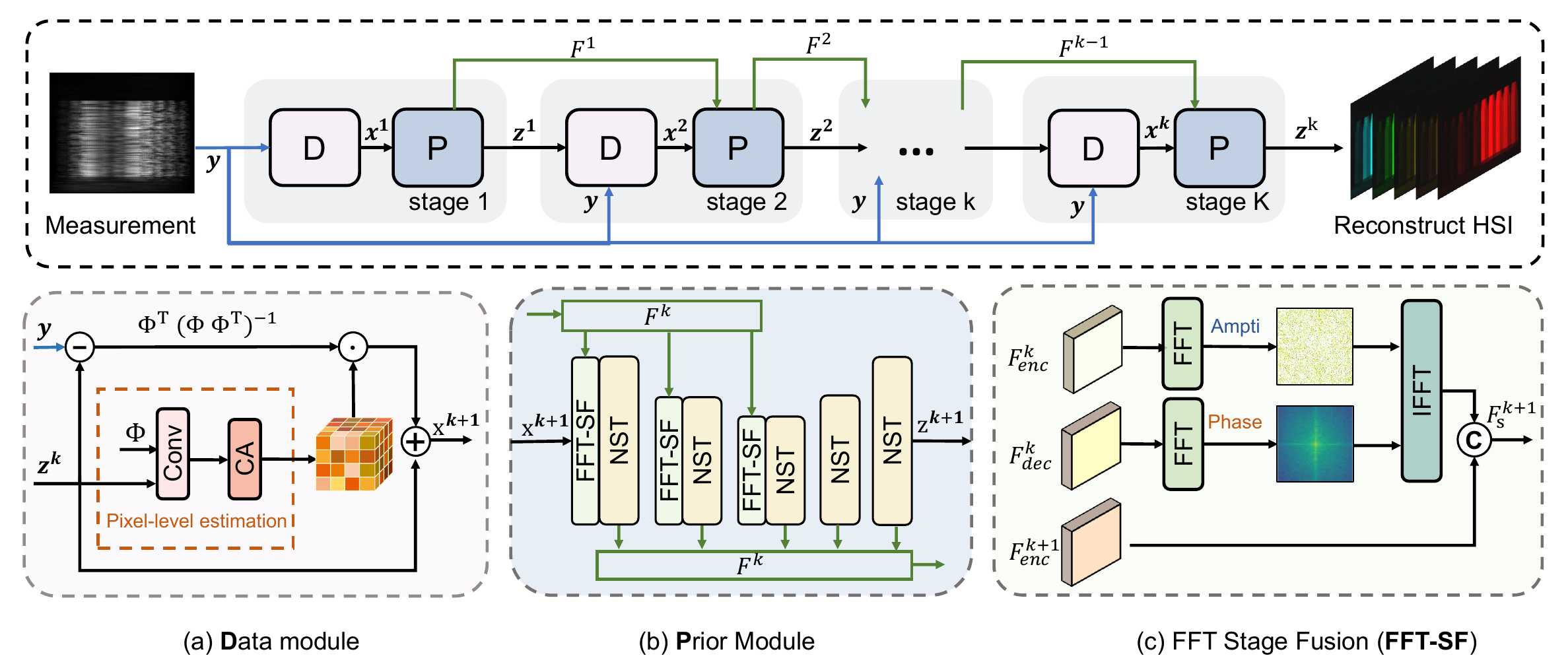}
\vspace{-2mm}                                  
\caption{Illustration of our proposed Pixel Adaptive Deep Unfolding Transformer (PADUT) for HSI reconstruction. Top: the overall architecture that consists of $K$ stages, each of which consists of a data module and a prior module. (a) Pixel-adaptive data module (b) Prior module (c) Stage fusion module}
\label{fig:overall}
\end{figure*}
\subsection{Problem Formulation}
Based on the compressive theory~\cite{donoho2006compressed}, the CASSI system can capture a compressed measurement that includes information covering all bands. Figure \ref{fig:cassi} illustrates a basic pipeline of the coding procedure. 
Considering a spectral image ${\rm\mathbf{X}_{\lambda}} \in \mathbb{R}^{M\times N}$ with $\lambda$ as its wavelength, 
the captured HSI from real scenes is firstly modulated via a coded aperture  ${\rm\mathbf{C}}_\lambda \in \mathbb{R}^{M\times N }$. The temporary measurement ${\rm\mathbf{Y}}_\lambda\in \mathbb{R}^{M\times N}$ is denoted as:
\begin{equation}
        {\rm \mathbf{Y}_\lambda = \mathbf{C}_\lambda \odot \mathbf{X}_\lambda},
\end{equation}
 where $\odot$ denotes the element-wise multiplication. 
 
 By shifting $\mathbf{Y}_\lambda$ along the horizontal direction according to the dispersive function $d$, the intermediate measurement $\rm\mathbf{Y}_\lambda^{\prime} \in \mathbb{R}^{M\times (N+B-1)}$ is modulated to:
\begin{equation}
        {\rm\mathbf{Y}_\lambda^{\prime}}(h,w) = {\rm\mathbf{G}_\lambda^{\prime}}(h,w+{d(\lambda))},
\end{equation}
where $h$ and $w$ denote the spatial coordinates. $B$ is the number of bands in the desired 3D HSI. In the presence of noise $N\in$$ \mathbb{R}^{M\times (N+B-1) \times B}$, the final measurement $\rm{\mathbf{Y}}\in \mathbb{R}^{M\times (N+B-1) \times B}$ can be formulated as:
 
\begin{equation}
        {\rm \mathbf{Y}} = \sum_{\lambda=1}^{B}\rm{\mathbf{G}_\lambda^{\prime}} + \mathbf{N}.
\end{equation}

For the connivance, with a shifted version of coded aperture $\mathbf{C_\lambda}$ as $\mathbf{\Phi} $, the overall imaging model is formulated as:

\begin{equation}
        \rm \mathbf{y} = \mathbf{\Phi} \mathbf{x} + \mathbf{n}.
\end{equation}

CASSI system makes sacrifices the spatial information to obtain the spectral information. Consequently, the spatial intensity in the coded measurement $\mathbf{y}$ incorporates a combined representation of spatial and spectral
information. 
This suggests that pixels in different locations in the HSI may have different levels of compression. This motivates us to improve in the optimization process for the pixel-specific reconstruction.

%The degradation process from high-quality HSI $\rm{X}$$\in  \mathbb{R}^{ B\times H\times W}$ to degraded measurement $\rm{Y}$ $\in  \mathbb{R}^{ B\times h\times w}$cam summarised as 

%The HSI reconstruction process aims to recover the desired clean HSI $\mathbf{x}$ from degraded observation $\mathbf{y}$.  
%Different types of degradation function $\varPhi$ lead to different reconstruction tasks. In the HSI denoising task, $\varPhi$ is an identity matrix, and only random noise $\mathbf{n}$ is considered as a distortion. In the HSI compressive sensing task, $\varPhi$ is a projection function that encodes the 3D data cube into a 2D measurement.
\subsection{Revisting the Deep Unrolling Framework}
Mathematically, the optimization of HSI reconstruction could be modeled as:
\begin{equation}
\hat{\rm{\mathbf{x}}} =  \arg\min_{\rm{\mathbf{x}}}{\frac{1}{2}{||\rm{\mathbf{y}}-\rm{\Phi}\rm{\mathbf{x}}||_2}^2}+\eta J(\rm{\mathbf{x}}),
\label{eq:argmin}
\end{equation}
where $J(\rm{\mathbf{x}})$ denoted the regularizer term with parameter $\eta$.

The coding mask $\Phi$ reveals the spatial relation as well as the spectral relation between the coded measurement and desired 3D data. Recent deep unfolding works~\cite{wang2020dnu,zhang2022herosnet,cai2022degradation} have shown the great potential of combing $\Phi$ with deep networks. In the half quadratic splitting (HQS) algorithm, Eq. \eqref{eq:argmin} is formulated into subproblems through the introduction of an auxiliary variable $\rm\textbf{z}$ as:
\begin{equation}
(\hat{\rm{\mathbf{x}}},\hat{\rm{\mathbf{z}}})  =  \arg\min_{\rm{\mathbf{x,z}}}{\frac{1}{2}||\rm{\mathbf{y}}-\rm{\Phi}\rm{\mathbf{x}}||_2^2}+\eta J(\rm{\mathbf{z}}) + \frac{\mu}{2}{{|| \mathbf{z}-\mathbf{x}||}_2^2},
\label{eq:zx}
\end{equation}
where $\mu$ is the penalty parameter. HQS algorithm approximatively optimizes Eq. \eqref{eq:zx} through two iterative convergence subproblems:
\begin{align}
\rm{{\mathbf x}^{k+1}} &= \arg\min_{\rm{\mathbf{x}}}{||{\rm\mathbf y}-\rm{\Phi}\rm{\mathbf{x^{k}}}||_2^2} + \mu {|| {\rm\mathbf z^{k}}-{\rm\mathbf x^{k}}||}_2^2 , \\
\rm{\mathbf{z}^{k+1}}& = \arg\min_{\rm{\mathbf{z}}}{\frac{\mu}{2}||\rm{\mathbf{z^{k}}}-\mathbf{x}^{k+1}||_2^2}  +\eta J(\rm{\mathbf{z^{k}}}),
\label{eq:opt_x}
\end{align}

In short, a close-form solution of $\rm\mathbf{x}$ is formulated by:
\begin{align}
\rm{\mathbf{x}^{k+1}} &= (\mathbf{\Phi}^{T}\mathbf{\Phi}+\mu {\rm\mathbf I})^{-1}(\mathbf{\Phi} \mathbf{y} + \mu \mathbf{z}^k) \\
& =
\mathbf{z}^{k} + \frac{1}{1+\mu} \mathbf{\Phi}^{T} ( \mathbf{\Phi\Phi}^{T})^{-1} (\mathbf{y}- \mathbf{\Phi} {\mathbf{z}^{k}}).
\label{eq:data}
\end{align}

In deep unrolling methods, $\mathbf{z}$ is often solved by:% successive deep networks via:
\begin{equation}
{\rm\textbf{z}}^{k+1} = {\rm P}_{k+1}({\rm\mathbf{x}}^{k+1}),
\label{eq:prior}
\end{equation}
where ${\rm P}_{k+1}$ refers to the deep network in the $k+1$ stage.

Usually, a deep unfolding framework consists of multiple stages specifically devised to reconstruct the underlying HSI cube $\rm\mathbf{x}$ from coded measurement $\rm\mathbf{y}$. Eq. \eqref{eq:data} serves as a data module that introduces the physical characteristics into optimization. Meanwhile, deep characteristics are exploited in Eq. \eqref{eq:prior} and can be referred to the prior module. 
%Nevertheless, existing deep unfoldingworks mainly focused on one specific reconstruction task.  
%In addition to the problem of insufficient generalization, the existing deep unfoldingworks treat the information interaction between stages as a simple  fusion problem.  There are actually some differences in the intrinsic characteristics of the recovered images at different stages.

As mentioned in the problem formulation, pixels in the HSI suffer varying degrees of information loss in the compressive sensing. Although the physical mask $\bm{\mathbf{\Phi}}$ alleviates such a problem in the prior module, it often takes a fixed way of assistance. Moreover, in a real sensing system, there is often a gap between the mask and the real degradation.

\subsection{ Framework}
Based on the aforementioned observation, we design a pixel-adaptive deep unfolding transformer for HSI reconstruction. Figure \ref{fig:overall} illustrates the general framework of our proposed approach, which is composed of $K$ stages to reconstruct a compressed HSI. In each stage, a data module is followed by a denoiser, which refers to the prior module. The data module aims to utilize the physical degradation information while the prior module is for optimization. Our denoiser is a U-shaped design. In the encoder, each layer contains a Fast Fourier Transformer stage fusion (FFT-SF) layer and a Non-local Spectral Transformer (NST) layer. The decoder is only composed of NST layers.

\noindent\textbf{Pixel-Adaptive Prior Module.}
Observing from Eq. \eqref{eq:data}, $\frac{1}{1+\mu} $ plays an important role in the optimization of $\rm \mathbf{x}$. For simplify, we use $\rm\mathbf{F_\sigma}$ to represent 
$\frac{1}{1+\mu}$ as:
\begin{equation}
\rm{\mathbf{x}^{k+1}} =
\rm{\mathbf z}^{k} + {\rm\mathbf{F}}_\sigma \mathbf{\Phi}^{T} ( \mathbf{\Phi\Phi}^{T})^{-1} (\mathbf{y}- \mathbf{\Phi} {\rm \mathbf{z}}^{k}).
\label{eq:adaptive_eq}
\end{equation}

%Moreover, for different reconstruction task, the degradation operation $\Phi$ reveals the intrinsic properties of the degradation process, likewise, provides guidance for the reconstruction. Therefore, we utilize $\Phi$ to generate the task-oriented trade-off parameter $\sigma$. For the denoising task, we use identity matrix $I$ as $\Phi$. 
In the compressive sensing process, patterns in different positions and bands are markedly different
 due to the modulation. Due to the presence of instrument noise, the distribution of noise is also varied in the HSI cube. This difference persists throughout the recovery process. Considering the problem of inconsistent and agnostic degradation at different locations in the HSI, we design a pixel-adaptive data module for the deep unfolding framework.
\begin{figure*}[t]
\scriptsize 
\centering	
\includegraphics[width=0.85\linewidth]{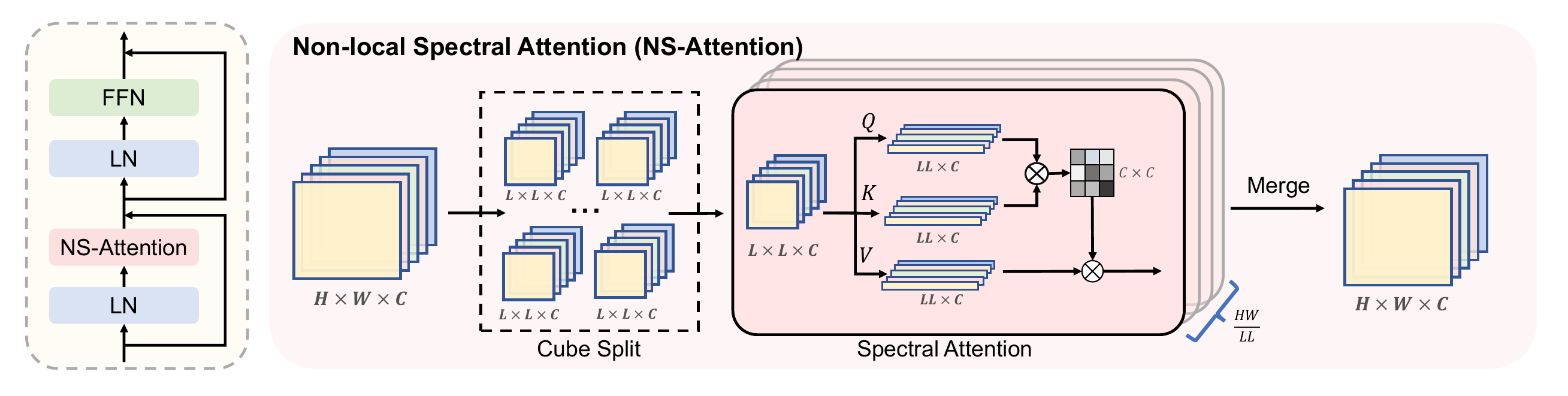}
\vspace{-2mm}                                  
\caption{Illustration of our Non-local Spectral Transformer (NST) layer. The core modules of our block are: LayerNorm (LN), feed-forward
network (FFN) and NS-Attention module.}
\label{fig:attention}
\end{figure*}

The details of our pixel-adaptive prior module are illustrated in Figure \ref{fig:overall} \textcolor{red}{(a)}. Since the physical mask $\mathbf{\rm\Phi}$ establishes a relevance of the spatial and spectral dimensions, and $\rm{\mathbf{z}^{k}}$ indicates the current input feature, we generate the 3D parameters $\rm \mathbf{F}_\sigma$ via the convolution layer and Channel Attention (CA) ~\cite{hu2018squeeze} layer. 
\begin{align}
    &\rm \mathbf{F}_{\sigma}^{\prime} = \rm{Conv} (\rm{Concat}[\rm{\mathbf z}^{k}, {\rm\mathbf\Phi} ]),\\
   & \rm \mathbf{F}_\sigma = \rm{CA}(\rm \mathbf{F}_{\sigma}^\prime),
 \end{align}
 
Then, the obtained 3D parameters $\rm \mathbf{F}_\sigma $ are used to parameterize the 3D data in eq. \eqref{eq:adaptive_eq} by the pixel-adaptive gradient descent step, achieving the pixel specific reconstruction.

\noindent\textbf{Non-local Spectral Transformer.}
The spectral self-attention~\cite{cai2022mask,zamir2022restormer} has shown a promising result in the image restoration area. However, it can hardly model the fine-grained similarity characteristic between pixels in both spatial and spectral dimensions. On the one hand, since spectral self-attention takes the pixels of the entire spectral dimension as the feature value to represent the spectral characteristic, local detail information can easily be lost. On the other hand, due to the coding property of the CASSI system, compressed information can often be found in adjacent areas. The spectral self-attention needs to better adapt to the 3D HSI cube and the coding system.

To make use of spatial-spectral information of HSI, we propose the Non-local Spectral Transformer (NST) for HSI restoration in the prior module.
As shown in Figure \ref{fig:attention}, NST consists of Layer Normalization (LN), a Non-local Spectral Attention (NSA), and a Feed-Forward
Network (FFN). The spatial shift operation is conducted between two NST to explore more than local features.

For the non-local spectral attention layer, we first split the entire feature of $\rm\mathbf{x}_{in}\in \mathbb{R}^{ H\times W\times C}$ into several cube patches as $\left\{\rm\mathbf{x}_{1},\rm\mathbf{x}_{2},...,\rm\mathbf{x}_{G}\right\}$. Each cube is of size $L$$\times$$L$$\times$$C$. For each cube, we project ${\rm\mathbf x}_{i}\in \mathbb{R}^{ LL\times C}$ into query ${\rm\mathbf{Q}}_{i} \in \mathbb{R}^{ LL\times C}$, key ${\rm\mathbf{K}}_i \in \mathcal{R}^{LL\times C}$, and value ${\rm\mathbf{Q}}_i \in \mathbf{R}^{LL\times C}$ as
\begin{align}
{\rm\mathbf Q}_i={\rm\mathbf x}_i{\rm \mathbf W}^{\rm Q},{\rm \mathbf K}_i={\rm\mathbf x}_i{\rm\mathbf W}^{\mathbf K},{\rm\mathbf V}_i={\rm\mathbf x}_i{\rm\mathbf W}^{{\mathbf Q}}.
\end{align}

After the projection, the self-attention features for each cube are calculated as:
\begin{align} 
{\rm Attention}({\rm\mathbf Q}_i,{\rm\mathbf K}_i,{\rm\mathbf V}_i) ={\rm\mathbf V}_i \rm{Softmax} ( \frac{{\rm\mathbf{K}_i}^T{{\rm\mathbf{Q}_i}}}{\beta}),
\end{align}
where the obtained attention map is of size $\mathbb{R}^{ C\times  C}$ for each spatial-spectral cube, capturing and consolidating the non-local information across the entire data volume. In the implementation, we adopt a similar approach of 
multi-head self-attention and partition the number of spectral bands into 'heads' and subsequently learn individual features.

\noindent\textbf{Fast Fourier Transform Stage Fusion.} Deep unfolding framework has shown the effectiveness of multi-stage learning via interpretable networks. Since the contextual information and detailed information varied at different stages, effectively employing the rich features could boost the performance of reconstruction \cite{zhang2022herosnet,mou2022deep}. Moreover, inside each stage, the encoder-decoder denoiser leads to contextually different intermediate features due to the inherent trade-off between spatial and spectral information. How to interpolate cross-stage features and inner-stage features more effectively remains an ongoing challenge.

As shown in Figure \ref{fig:fft}, in the frequency domain, the phase component and amplitude component of recovery HSI in different stages 
correspond differently. In the encoder, the magnitude information is more prominent. In the later decoder, the phase information is more clear. According to this observation, we introduce the Fast Fourier Transform to the inter-stage connection to obtain a better reconstruction result from the frequency domain. 

The details of our FFT-SF is shown in Figure \ref{fig:overall} \textcolor{red}{(c)}. We first transform the encoder and decoder feature from the former layer into Fourier domain. Then, a Fourier-based fusion is conducted to focus on the different frequency characteristics. Last, the frequency-enhanced feature is used to enhance the feature of next stage.

To model the frequency feature of $\rm \mathbf{x}$ with a shape of $\mathbb{R}^{H \times W\times C}$, we leverage the Fourier transform $\mathcal{F}$ to convert it into Fourier domain, which is formulated as $ \mathcal{F}(\rm\mathbf{x})$:
\begin{align}
    \mathcal{F}(\rm\mathbf{x}) &= {\rm\mathbf{x}}(u,v) \\%= X(u,v) \\
    &=\frac{1}{\sqrt{HW}}\sum_{h=0}^{H-1}\sum_{w=0}^{W-1}{\rm\mathbf{x}}(h,w)e^{-j2\pi{(\frac{h}{H}u+\frac{w}{W}v)}},
\end{align}
where $u$ and $v$ stand for coordinates in frequency domain.

\begin{table*}[htbp]
  \centering

    \resizebox{\textwidth}{!}{
  \setlength{\tabcolsep}{2mm}
  	\renewcommand{\arraystretch}{0.8}
    \begin{tabular}{cccccccccccccc}
    \midrule[0.8pt]
        \rowcolor[rgb]{ .949,  .949,  .949} & Params & GFLOPs & \multicolumn{1}{c}{s1} & \multicolumn{1}{c}{s2} & \multicolumn{1}{c}{s3} & \multicolumn{1}{c}{s4} & \multicolumn{1}{c}{s5} & \multicolumn{1}{c}{s6} & \multicolumn{1}{c}{s7} & \multicolumn{1}{c}{s8} & \multicolumn{1}{c}{s9} & \multicolumn{1}{c}{s10} & \multicolumn{1}{c}{Avg} \bigstrut\\
    \midrule[0.8pt]
    \multirow{2}[2]{*}{TwIST~\cite{bioucas2007new}} & \multirow{2}[2]{*}{-} & \multirow{2}[2]{*}{-} & \multicolumn{1}{c}{25.16} & \multicolumn{1}{c}{23.02} & \multicolumn{1}{c}{21.40} & \multicolumn{1}{c}{30.19} & \multicolumn{1}{c}{21.41} & \multicolumn{1}{c}{20.95} & \multicolumn{1}{c}{22.20} & \multicolumn{1}{c}{21.82} & \multicolumn{1}{c}{22.42} & \multicolumn{1}{c}{22.67} & \multicolumn{1}{c}{23.12} \bigstrut[t]\\
          &       &       & \multicolumn{1}{c}{0.700} & \multicolumn{1}{c}{0.604} & \multicolumn{1}{c}{0.711} & \multicolumn{1}{c}{0.851} & \multicolumn{1}{c}{0.635} & \multicolumn{1}{c}{0.644} & \multicolumn{1}{c}{0.643} & \multicolumn{1}{c}{0.650} & \multicolumn{1}{c}{0.690} & \multicolumn{1}{c}{0.569} & \multicolumn{1}{c}{0.669} \bigstrut[b]\\
    \hline
    \multirow{2}[2]{*}{GAP-TV~\cite{yuan2016generalized}} & \multirow{2}[2]{*}{-} & \multirow{2}[2]{*}{-} & \multicolumn{1}{c}{26.82} & \multicolumn{1}{c}{22.89} & \multicolumn{1}{c}{26.31} & \multicolumn{1}{c}{30.65} & \multicolumn{1}{c}{23.64} & \multicolumn{1}{c}{21.85} & \multicolumn{1}{c}{23.76} & \multicolumn{1}{c}{21.98} & \multicolumn{1}{c}{22.63} & \multicolumn{1}{c}{23.1} & \multicolumn{1}{c}{24.36} \bigstrut[t]\\
          &       &       & \multicolumn{1}{c}{0.754} & \multicolumn{1}{c}{0.610} & \multicolumn{1}{c}{0.802} & \multicolumn{1}{c}{0.852} & \multicolumn{1}{c}{0.703} & \multicolumn{1}{c}{0.663} & \multicolumn{1}{c}{0.688} & \multicolumn{1}{c}{0.655} & \multicolumn{1}{c}{0.682} & \multicolumn{1}{c}{0.584} & \multicolumn{1}{c}{0.669} \bigstrut[b]\\
    \hline
    \multirow{2}[2]{*}{DeSCI~\cite{liu2018rank}} & \multirow{2}[2]{*}{-} & \multirow{2}[2]{*}{-} & \multicolumn{1}{c}{27.13} & \multicolumn{1}{c}{23.04} & \multicolumn{1}{c}{26.62} & \multicolumn{1}{c}{34.96} & \multicolumn{1}{c}{23.94} & \multicolumn{1}{c}{22.38} & \multicolumn{1}{c}{24.45} & \multicolumn{1}{c}{22.03} & \multicolumn{1}{c}{24.56} & \multicolumn{1}{c}{23.59} & \multicolumn{1}{c}{25.27} \bigstrut[t]\\
          &       &       & \multicolumn{1}{c}{0.748} & \multicolumn{1}{c}{0.620} & \multicolumn{1}{c}{0.818} & \multicolumn{1}{c}{0.897} & \multicolumn{1}{c}{0.706} & \multicolumn{1}{c}{0.683} & \multicolumn{1}{c}{0.743} & \multicolumn{1}{c}{0.673} & \multicolumn{1}{c}{0.732} & \multicolumn{1}{c}{0.587} & \multicolumn{1}{c}{0.721} \bigstrut[b]\\
    \hline
    \multirow{2}[2]{*}{$\lambda$-Net~\cite{miao2019net}} & \multirow{2}[2]{*}{62.64M } & \multirow{2}[2]{*}{117.98} & \multicolumn{1}{c}{30.10} & \multicolumn{1}{c}{28.49} & \multicolumn{1}{c}{27.73} & \multicolumn{1}{c}{37.01} & \multicolumn{1}{c}{26.19} & \multicolumn{1}{c}{28.64} & \multicolumn{1}{c}{26.47} & \multicolumn{1}{c}{26.09} & \multicolumn{1}{c}{27.50} & \multicolumn{1}{c}{27.13} & \multicolumn{1}{c}{28.53} \bigstrut[t]\\
          &       &       & \multicolumn{1}{c}{0.849} & \multicolumn{1}{c}{0.805} & \multicolumn{1}{c}{0.870} & \multicolumn{1}{c}{0.934} & \multicolumn{1}{c}{0.817} & \multicolumn{1}{c}{0.853} & \multicolumn{1}{c}{0.806} & \multicolumn{1}{c}{0.831} & \multicolumn{1}{c}{0.826} & \multicolumn{1}{c}{0.816} & \multicolumn{1}{c}{0.841} \bigstrut[b]\\
    \hline
    \multirow{2}[2]{*}{TSA-Net~\cite{meng2020end}} & \multirow{2}[2]{*}{44.25M } & \multirow{2}[2]{*}{110.06} & \multicolumn{1}{c}{32.03} & \multicolumn{1}{c}{31.00} & \multicolumn{1}{c}{32.25} & \multicolumn{1}{c}{39.19} & \multicolumn{1}{c}{29.39} & \multicolumn{1}{c}{31.44} & \multicolumn{1}{c}{30.32} & \multicolumn{1}{c}{29.35} & \multicolumn{1}{c}{30.01} & \multicolumn{1}{c}{29.59} & \multicolumn{1}{c}{31.46} \bigstrut[t]\\
          &       &       & \multicolumn{1}{c}{0.892} & \multicolumn{1}{c}{0.858} & \multicolumn{1}{c}{0.915} & \multicolumn{1}{c}{0.953} & \multicolumn{1}{c}{0.884} & \multicolumn{1}{c}{0.908} & \multicolumn{1}{c}{0.878} & \multicolumn{1}{c}{0.888} & \multicolumn{1}{c}{0.890} & \multicolumn{1}{c}{0.874} & \multicolumn{1}{c}{0.894} \bigstrut[b]\\
    \hline
    \multirow{2}[2]{*}{DGSMP~\cite{huang2021deep}} & \multirow{2}[2]{*}{3.76M } & \multicolumn{1}{c}{\multirow{2}[2]{*}{646.65}} & \multicolumn{1}{c}{33.26} & \multicolumn{1}{c}{32.09} & \multicolumn{1}{c}{33.06} & \multicolumn{1}{c}{40.54} & \multicolumn{1}{c}{28.86} & \multicolumn{1}{c}{33.08} & \multicolumn{1}{c}{30.74} & \multicolumn{1}{c}{31.55} & \multicolumn{1}{c}{31 .66} & \multicolumn{1}{c}{31 .44} & \multicolumn{1}{c}{32.63} \bigstrut[t]\\
          &       &       & \multicolumn{1}{c}{0.915} & \multicolumn{1}{c}{0.898} & \multicolumn{1}{c}{0.925} & \multicolumn{1}{c}{0.964} & \multicolumn{1}{c}{0.882} & \multicolumn{1}{c}{0.937} & \multicolumn{1}{c}{0.886} & \multicolumn{1}{c}{0.923} & \multicolumn{1}{c}{0.911} & \multicolumn{1}{c}{0.925} & \multicolumn{1}{c}{0.917} \bigstrut[b]\\
    \hline
    \multirow{2}[2]{*}{GAP-Net~\cite{meng2020gap} } & \multirow{2}[2]{*}{4.27M } & \multicolumn{1}{c}{\multirow{2}[2]{*}{78.58}} & \multicolumn{1}{c}{33.74} & \multicolumn{1}{c}{33.26} & \multicolumn{1}{c}{34.28} & \multicolumn{1}{c}{41.03} & \multicolumn{1}{c}{31.44} & \multicolumn{1}{c}{32.40} & \multicolumn{1}{c}{32.27} & \multicolumn{1}{c}{30.46} & \multicolumn{1}{c}{33.51} & \multicolumn{1}{c}{30.24} & \multicolumn{1}{c}{33.26} \bigstrut[t]\\
          &       &       & \multicolumn{1}{c}{0.911} & \multicolumn{1}{c}{0.900} & \multicolumn{1}{c}{0.929} & \multicolumn{1}{c}{0.967} & \multicolumn{1}{c}{0.919} & \multicolumn{1}{c}{0.925} & \multicolumn{1}{c}{0.902} & \multicolumn{1}{c}{0.905} & \multicolumn{1}{c}{0.915} & \multicolumn{1}{c}{0.895} & \multicolumn{1}{c}{0.917} \bigstrut[b]\\
    \hline
    \multirow{2}[2]{*}{HDNet~\cite{hu2022hdnet}} & \multirow{2}[2]{*}{2.37M } & \multicolumn{1}{c}{\multirow{2}[2]{*}{154.76}} & \multicolumn{1}{c}{35.14} & \multicolumn{1}{c}{35.67} & \multicolumn{1}{c}{36.03} & \multicolumn{1}{c}{42.30} & \multicolumn{1}{c}{32.69} & \multicolumn{1}{c}{34.46} & \multicolumn{1}{c}{33.67} & \multicolumn{1}{c}{32.48} & \multicolumn{1}{c}{34.89} & \multicolumn{1}{c}{32.38} & \multicolumn{1}{c}{34.97} \bigstrut[t]\\
          &       &       & \multicolumn{1}{c}{0.935} & \multicolumn{1}{c}{0.940} & \multicolumn{1}{c}{0.943} & \multicolumn{1}{c}{0.969} & \multicolumn{1}{c}{0.946} & \multicolumn{1}{c}{0.952} & \multicolumn{1}{c}{0.926} & \multicolumn{1}{c}{0.941} & \multicolumn{1}{c}{0.942} & \multicolumn{1}{c}{0.937} & \multicolumn{1}{c}{0.943} \bigstrut[b]\\
    \hline
    \multirow{2}[2]{*}{MST-L~\cite{cai2022mask}} & \multirow{2}[2]{*}{2.03M} & \multicolumn{1}{c}{\multirow{2}[2]{*}{28.15}} & \multicolumn{1}{c}{35.40} & \multicolumn{1}{c}{35.87} & \multicolumn{1}{c}{36.51} & \multicolumn{1}{c}{42.27} & \multicolumn{1}{c}{32.77} & \multicolumn{1}{c}{34.80} & \multicolumn{1}{c}{33.66} & \multicolumn{1}{c}{32.67} & \multicolumn{1}{c}{35.39} & \multicolumn{1}{c}{32.50} & \multicolumn{1}{c}{35.18} \bigstrut[t]\\
          &       &       & \multicolumn{1}{c}{0.941} & \multicolumn{1}{c}{0.944} & \multicolumn{1}{c}{0.953} & \multicolumn{1}{c}{0.973} & \multicolumn{1}{c}{0.947} & \multicolumn{1}{c}{0.955} & \multicolumn{1}{c}{0.925} & \multicolumn{1}{c}{0.948} & \multicolumn{1}{c}{0.949} & \multicolumn{1}{c}{0.941} & \multicolumn{1}{c}{0.948} \bigstrut[b]\\
    \hline
    \multirow{2}[2]{*}{CST-L~\cite{yun2022coarse}} & \multirow{2}[2]{*}{3.00M } & \multicolumn{1}{c}{\multirow{2}[2]{*}{40.01}} & \multicolumn{1}{c}{35.96} & \multicolumn{1}{c}{36.84} & \multicolumn{1}{c}{38.16} & \multicolumn{1}{c}{42.44} & \multicolumn{1}{c}{33.25} & \multicolumn{1}{c}{35.72} & \multicolumn{1}{c}{34.86} & \multicolumn{1}{c}{34.34} & \multicolumn{1}{c}{36.51} & \multicolumn{1}{c}{33.09} & \multicolumn{1}{c}{36.12} \bigstrut[t]\\
          &       &       & \multicolumn{1}{c}{0.949} & \multicolumn{1}{c}{0.955} & \multicolumn{1}{c}{0.962} & \multicolumn{1}{c}{0.975} & \multicolumn{1}{c}{0.955} & \multicolumn{1}{c}{0.963} & \multicolumn{1}{c}{0.944} & \multicolumn{1}{c}{0.961} & \multicolumn{1}{c}{0.957} & \multicolumn{1}{c}{0.945} & \multicolumn{1}{c}{0.957} \bigstrut[b]\\
    \hline
    \multirow{2}[2]{*}{DAUHST-L~\cite{cai2022degradation}} & \multirow{2}[2]{*}{6.15M } & \multirow{2}[2]{*}{79.50} & \multicolumn{1}{c}{37.25} & \multicolumn{1}{c}{39.02} & \multicolumn{1}{c}{41.05} & \multicolumn{1}{c}{46.15} & \multicolumn{1}{c}{35.80} & \multicolumn{1}{c}{37.08} & \multicolumn{1}{c}{37.57} & \multicolumn{1}{c}{35.10} & \multicolumn{1}{c}{40.02} & \multicolumn{1}{c}{\textbf{34.59}} & \multicolumn{1}{c}{38.36} \bigstrut[t]\\
          &       &       & \multicolumn{1}{c}{0.958} & \multicolumn{1}{c}{0.967} & \multicolumn{1}{c}{0.971} & \multicolumn{1}{c}{0.983} & \multicolumn{1}{c}{0.969} & \multicolumn{1}{c}{0.970} & \multicolumn{1}{c}{0.963} & \multicolumn{1}{c}{0.966} & \multicolumn{1}{c}{0.970} & \multicolumn{1}{c}{0.956} & \multicolumn{1}{c}{0.967} \bigstrut[b]\\
    \hline
    % \multirow{2}[2]{*}{RDLUF-MixS2-L} & \multirow{2}[2]{*}{1.89M} & \multirow{2}[2]{*}{115.34} & \multicolumn{1}{c}{37.94} & \multicolumn{1}{c}{40.95} & \multicolumn{1}{c}{43.25} & \multicolumn{1}{c}{47.83} & \multicolumn{1}{c}{37.11} & \multicolumn{1}{c}{37.47} & \multicolumn{1}{c}{38.58} & \multicolumn{1}{c}{35.5} & \multicolumn{1}{c}{41.83} & \multicolumn{1}{c}{35.23} & \multicolumn{1}{c}{39.57} \bigstrut[t]\\
    %       &       &       & \multicolumn{1}{c}{0.966} & \multicolumn{1}{c}{0.977} & \multicolumn{1}{c}{0.979} & \multicolumn{1}{c}{0.99} & \multicolumn{1}{c}{0.976} & \multicolumn{1}{c}{0.975} & \multicolumn{1}{c}{39.69} & \multicolumn{1}{c}{0.97} & \multicolumn{1}{c}{0.978} & \multicolumn{1}{c}{0.962} & \multicolumn{1}{c|}{0.974} \bigstrut[b]\\
     \rowcolor[rgb]{ .996,  .973,  .957} & & &  36.25 & 37.92 & 39.63 & 44.55 & 34.59 & 35.58 & 35.69 & 33.76 & 38.26 & 33.24     & 36.95 \bigstrut[t]\\
      \rowcolor[rgb]{ .996,  .973,  .957}   \multirow{-2}[2]{*}{PADUT-3stg} & \multirow{-2}[2]{*}{1.35M} & \multirow{-2}[2]{*}{22.91} &  
    0.951 & 0.963 & 0.970 & 0.985 & 0.964 & 0.965 & 0.950 & 0.960 & 0.963 & 0.947 &    0.962  \bigstrut[b]\\
    \hline

    \rowcolor[rgb]{ .996,  .973,  .957} & & &     36.68 & 38.74 & 41.37 & 45.79 & 35.13 & 36.37 & 36.52 & 34.40 & 39.57 & 33.78 & 37.84 \bigstrut[t]\\
        \rowcolor[rgb]{ .996,  .973,  .957}   \multirow{-2}[2]{*}{PADUT-5stg} & \multirow{-2}[2]{*}{2.24M} & \multirow{-2}[2]{*}{37.90} &   0.955 & 0.969 & 0.975 & 0.988 & 0.967 & 0.969 & 0.959 & 0.967 & 0.971 & 0.955 & 0.967 \bigstrut\\
    \hline
     \rowcolor[rgb]{ .996,  .973,  .957}  & & &       37.34 & 39.74 & 41.92 & \textbf{47.01} & 35.70 & 36.73 & 37.01 & 34.68 & 39.51 & 34.43 
 &
    38.41\bigstrut[t]\\
       \rowcolor[rgb]{ .996,  .973,  .957}    \multirow{-2}[2]{*}{PADUT-7stg} & \multirow{-2}[2]{*}{3.14M} & \multirow{-2}[2]{*}{52.90} &        0.961 & 0.974 & 0.976 & 0.990 & 0.971 & 0.972 & 0.960 & 0.970 & 0.972 & 0.961  &	0.971  \bigstrut[b]\\
          \hline
    \rowcolor[rgb]{ .996,  .973,  .957}          &   & &   \textbf{ 37.36} & \textbf{40.43} & \textbf{42.38} & 46.62 & \textbf{36.26} & \textbf{37.27} & \textbf{37.83} & \textbf{35.33} & \textbf{40.86} & 34.55  & \textbf{38.89} \bigstrut[t]\\
    \rowcolor[rgb]{ .996,  .973,  .957}    \multirow{-2}[2]{*}{PADUT-12stg} & \multirow{-2}[2]{*}{5.38M} & \multirow{-2}[2]{*}{90.46}     &   \textbf{0.962} & \textbf{0.978} & \textbf{0.979} & \textbf{0.990} & \textbf{0.974} & \textbf{0.974} & \textbf{0.966} & \textbf{0.974} & \textbf{0.978} & \textbf{0.963}
        &  \textbf{0.974} \bigstrut[b]\\
   \midrule[0.8pt]
    \end{tabular}%
    }
\vspace{-2mm}    
      \caption{Results on 10 simulated scenes on KAIST dataset (S1$\sim$S10). The best results are in bold.}
  \label{tab:cs_simu_cave}%
\end{table*}%

% Since an image or feature may
% contain multiple channels, we separately apply Fourier intuitivetransform to each channel
% in our work with the FFT transform.
To analyze and utilize the frequency  characteristics of HSIs, we decompose the complex component ${\rm\mathbf x}(u, v)$ into amplitude  $A({\rm\mathbf{x}})$ and phase $P({\rm \mathbf x})$. The amplitude component provides insight into the intensity of pixels, whereas phase component is critical for conveying positional information. Following \cite{huang2022deep}, the mathematical formulation is given by:
\begin{align}
A({\rm\mathbf{x}}(u,v) &= \sqrt{R^2({\rm\mathbf x}(u,v))+I^2({\rm\mathbf x}(u,v))}, \\
P({\rm\mathbf x}(u,v)) &= {\rm arctan}[\frac{I({{\rm\mathbf x}}(u,v)}{R({\rm\mathbf x}(u,v))}],
\end{align}
where $R({\rm\mathbf{x}})$ and $I(\rm\mathbf{x})$ present the real and imaginary parts. 

For the ($k$+1)-th stage, the feature from the former stage denotes as $F_{enc}^{k}$ and $F_{dec}^{k}$, feature from the current encoder layer as $F_{enc}^{k+1}$. Based on the observation that amplitude information and phase information are expressed differently in encoder and decoder, FFT stage fusion is expressed as:
\begin{align}
{F_{s}^{k+1}}^{\prime} = \mathcal{F}^{-1}(A(F_{enc}^{k}), P(F_{dec}^{k})),\\
F_{s}^{k+1} = {\rm Conv}({\rm Concat}( {F_{s}^{k+1}}^{\prime} ,  F_{enc}^{k+1})]),
\end{align}
where $\mathcal{F}^{-1}$ represents the inverse Fourier transform.

% Table generated by Excel2LaTeX from sheet '10-70'

\section{Experiments}
In this section, we initially introduce the experimental setup and implementation specifics. Subsequently, we assess the performance of our proposed method on both synthetic data and real-world HSI datasets. Lastly, we conduct an ablation study to showcase the efficacy of our approach.
\noindent \textbf{Datasets.} For the simulated experiment, we utilized two widely used HSI datasets, namely CAVE and KAIST, for training and testing. 
The KAIST dataset
consists of 30 HSIs with a spatial resolution at 2704$\times$3376 and a spectral dimension of 31. The CAVE dataset comprises 32 HSIs of size 512$\times$ 512 $\times$ 31. In accordance with the settings in  TSA-Net~\cite{meng2020end}, we adopt CAVE dataset as the training set and 10 scenes from KAIST dataset as testing set. The patch size of each HSI is $256\times$$ 256 $$\times$$ 28$. 

For the real data experiment, five real HSIs collected in TSA-Net \cite{meng2020end} are used for evaluation. Each testing sample is of size $660 $$\times$$ 660 $$\times$$28$. Following \cite{cai2022mask}, training samples are extracted from the CAVE dataset and KAIST dataset with the patch size of $660$$\times$$660$$\times 28$.

\begin{figure*}[t]
\scriptsize 
\centering	
\includegraphics[width=1\linewidth]{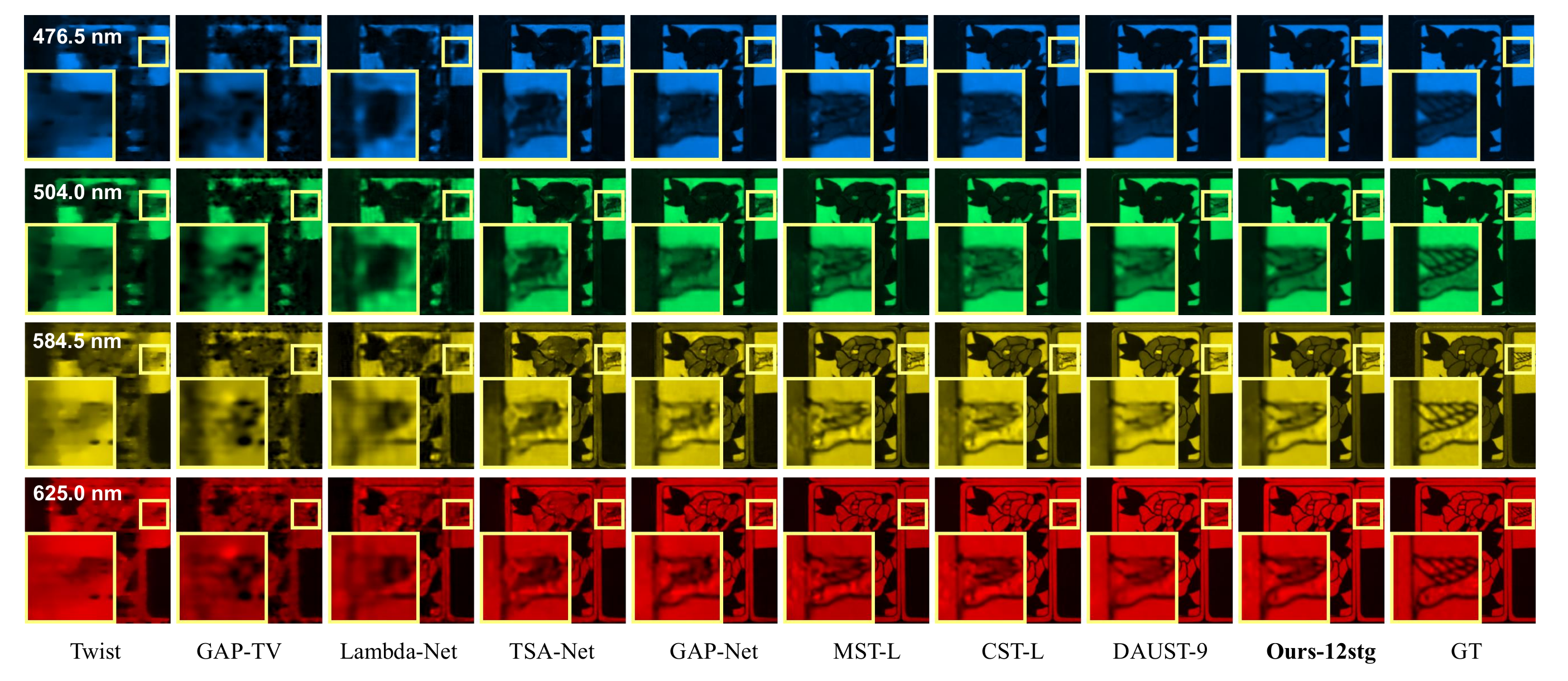}
\vspace{-7mm}
\caption{Visual comparisons on the KAIST dataset of scene 07 with  4 spectral channels.}
\label{fig:simu_resuls}
\end{figure*}
\noindent \textbf{Implementation.}
Our PADUT is implemented with Pytorch and trained with Adam~\cite{kingma2014adam} optimizer for 300 epochs. During training, the learning rate
is 4$\times$$10^{-4}$ using
the cosine annealing scheduler. The batch size is set to 5.

\noindent \textbf{Competing Methods.}
We compare our method with three classic model-based spectral reconstruction methods (TwIST~\cite{bioucas2007new}, GAP-TV~\cite{yuan2016generalized} and DeSCI~\cite{liu2018rank}), five end-to-end methods (Lambda-Net ~\cite{miao2019net}, TSA-Net~\cite{meng2020end},  MST~\cite{cai2022mask}, HDNet~\cite{hu2022hdnet} and CST\cite{yun2022coarse} ) and three deep unfolding methods (GAP-Net~\cite{meng2020gap},  DGSMP~\cite{huang2021deep}, and DAUHST~\cite{cai2022degradation}). 

\noindent \textbf{Evaluation Metrics.} The reconstruction quality is evaluated using peak signal-to-noise ratio (PSNR) and structural similarity index (SSIM).

\subsection{Simulation Results}
\begin{figure}[]
\scriptsize
\centering
%\small\scriptsize

  \includegraphics[width=0.95\linewidth]{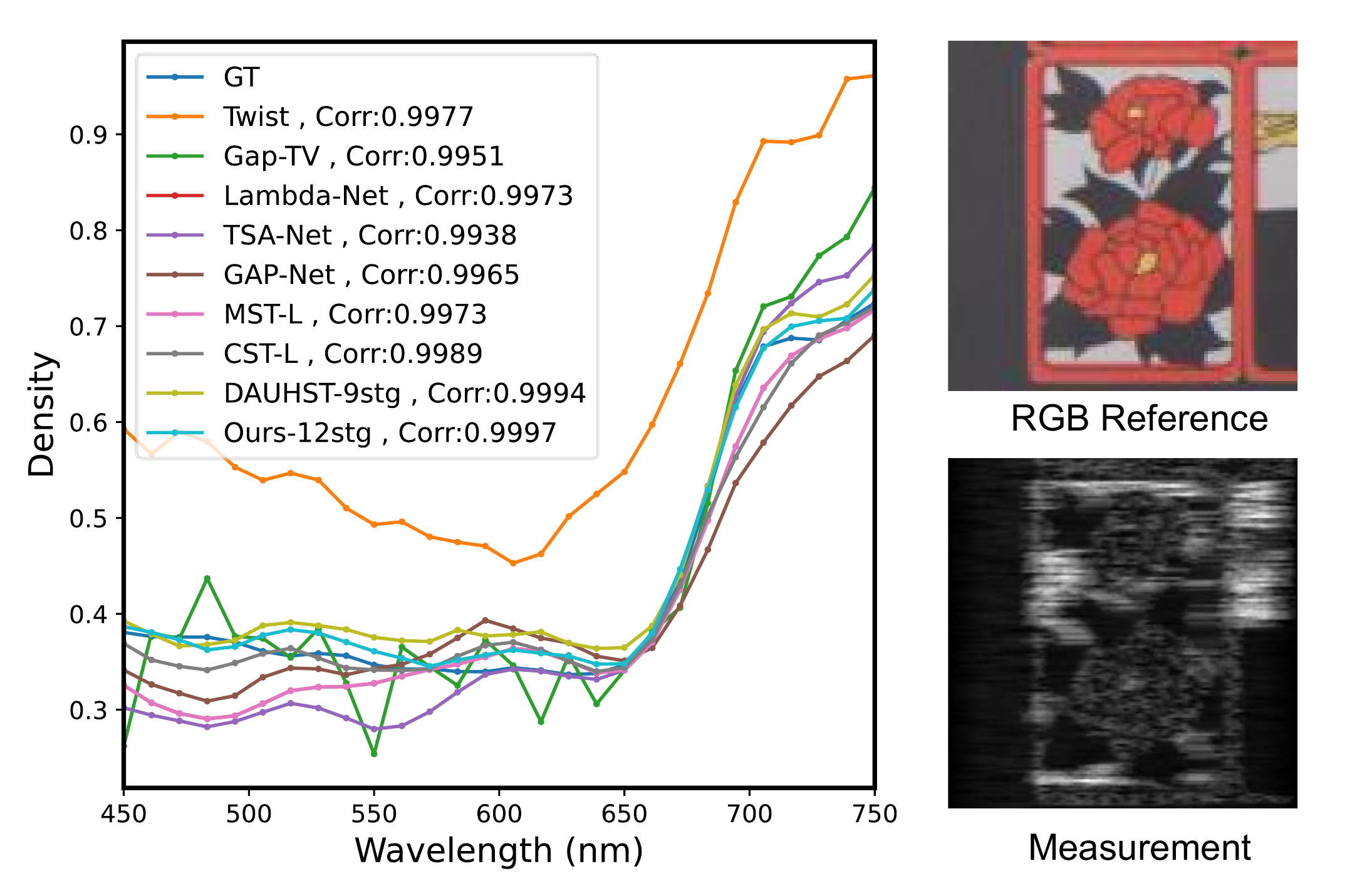} 
  \vspace{-3mm}
 \caption{ Spectral density curve on the simulation dataset of scene 07 with its corresponding RGB image and compressed measurement.}
 \vspace{-4mm}
 \label{fig:curve}
\end{figure}

\noindent \textbf{Numerical Results.}
The results from 10 simulated scenes are represented in Table \ref{tab:cs_simu_cave}. From the numerical results, we can observe that our method achieves the best result in almost all scenes and metrics, verifying the effectiveness of our method. Compared to end-to-end methods MST-L and CST-L, our PADUT-3stg achieves an improvement of 0.8 dB in PSNR. Our larger version PADUT-7stg outperforms DAUHST-L with a cost of 66.5$\%$ (52.9$/$79.5) GFLOPs. This highlights the benefit of leveraging the intrinsic characteristics of the CASSI system in pixel-level. Moreover, our method performs particularly well on the metric of SSIM. Figure \ref{fig:performance} reports the SSIM-GFLOPs comparisons of our method and recent HSI reconstruction methods.

\noindent \textbf{Visual Results.}
For better vision, following \cite{meng2020end}, we show the visual results in RGB format with CIE color as the mapping function. Figure \ref{fig:simu_resuls}
demonstrates that our method excels in preserving clearer spatial details, particularly in the zoomed area. Figure \ref{fig:curve} illustrates the corresponding spectral curves of competing methods as well as our method. The curve of our method is closest to the GroudTruth, indicating the spectral fidelity of our method.

\begin{figure*}[t]
\scriptsize 
\centering	

\includegraphics[width=1\linewidth]{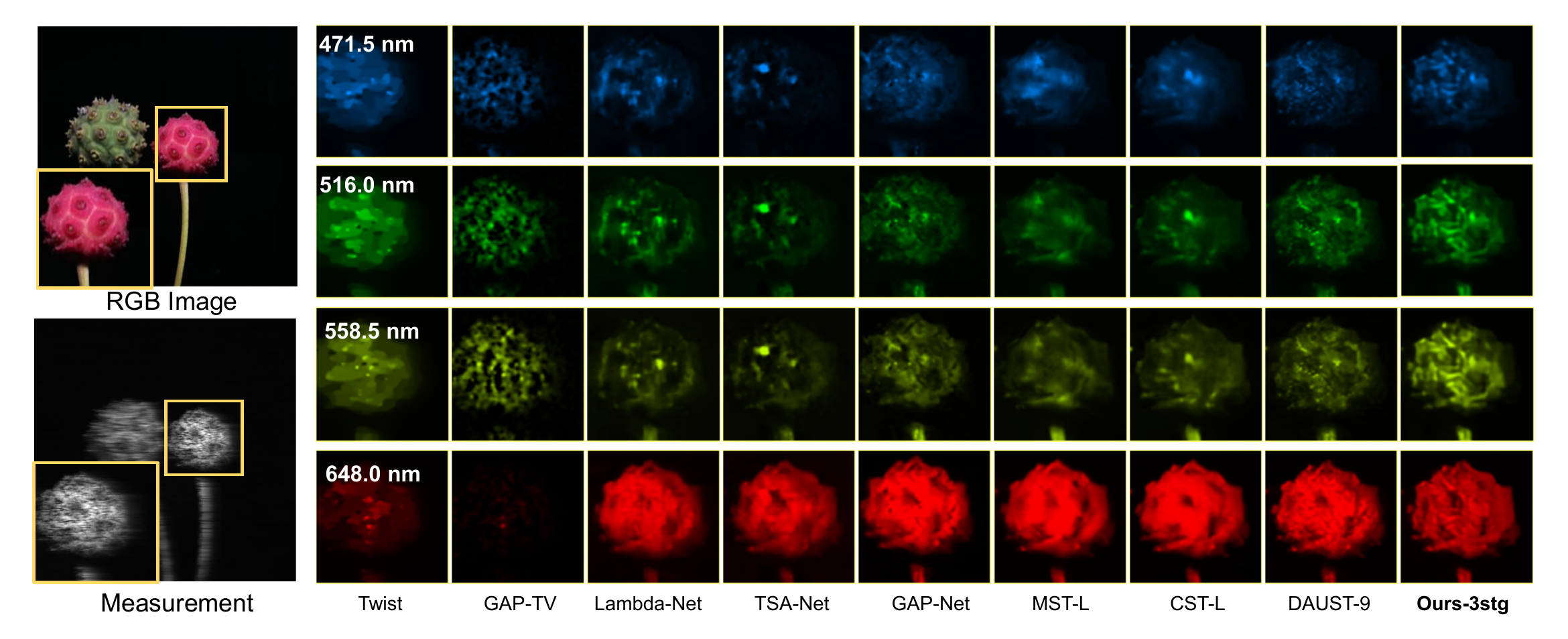}
\vspace{-6mm}
\caption{Reconstructed real HSI comparison   on scene 4 from the real dataset.}
\label{fig:real_results}

\end{figure*}
   \begin{table}[t]
  \centering
  \scriptsize
      \resizebox{\linewidth}{!}{
  \setlength{\tabcolsep}{1.5mm}
  	\renewcommand{\arraystretch}{0.8}
    \begin{tabular}{cccccccc}
    \midrule[0.8pt]
    \rowcolor[rgb]{ .949,  .949,  .949} Baseline & PA   & FFT-SF& NST & Params & GFLOPs & PSNR  & SSIM \bigstrut\\
    \midrule[0.8pt]
       \checkmark   &      &       &   &   1.30M	&20.31
    &         36.19 &0.959\bigstrut\\
    \hline
       \checkmark    &   \checkmark     &        &       &    1.33M   &   22.41    & 36.37      &  0.961                                                     \bigstrut\\
    \hline
        \checkmark   &    \checkmark    &    \checkmark  &      &      1.35M &   22.91    &   36.84      & 0.962 \bigstrut\\
    \hline
       \checkmark    &   \checkmark     &   \checkmark   &  \checkmark     &      1.35M   &  22.91     &  \textbf{36.95}     & \textbf{0.962} \bigstrut\\
    \midrule[0.8pt]
    \end{tabular}%
    }
    \vspace{-3mm}
      \caption{Break-down ablation study on individual components of the proposed method. }
      
  \label{tab:bre_ablation}%
\end{table}%

% \begin{table*}[]
% \begin{minipage}[t]{0.5\textwidth}
%  \centering

%         \resizebox{\linewidth}{!}{
%   \setlength{\tabcolsep}{5mm}
%   	\renewcommand{\arraystretch}{0.8}
%     \begin{tabular}{ccccc}
%     \midrule[0.8pt]
%      \rowcolor[rgb]{ .949,  .949,  .949}  & \multicolumn{1}{c}{Baseline } & \multicolumn{1}{c}{Concat} & \multicolumn{1}{c}{ISFF} & \multicolumn{1}{c}{FFT-SF (Ours)} \bigstrut\\
%        \midrule[0.8pt]
%         Params &      1.33M    &    1.47M   &    1.35M   &  1.35M \bigstrut\\
%     \hline
%     GFLOPs &    22.41
%   &   26.71  &   23.00    &  22.91 \bigstrut\\

%     \hline
%     PSNR  &    36.41  &   36.60    &  36.53     &  36.95 \bigstrut\\
%     \hline
%     SSIM  &   0.954  &  0.959     &   0.959      &  0.962\bigstrut\\
%     % \hline
%     % SAM   &       &       &       &  \bigstrut\\
%     \midrule[0.8pt]

%     \end{tabular}%
%     }
%     \vspace{-2mm}
% \caption{Ablation study of different stage fusion.}
%   \label{tab:stg_ablation}%   
%   \hspace{2mm}
% \end{minipage}
% \begin{minipage}[t]{0.48\textwidth}

%\end{minipage}
%\end{table*}

\subsection{Real Results}
The visual results of a real-scene  reconstructed HSI are
shown in Figure \ref{fig:real_results}. While most methods fail to reconstruct the detailed texture details, our method can successfully recover clear details. Compared to the deep unfolding-based approach DAUHST, we introduce the pixel-level reconstruction data module and FFT-based stage fusion, thus achieving better reconstruction results.

\subsection{Ablation Study}
To verify the effectiveness of our proposed structure,
we conduct the ablation study with Ours-3stg method. All the evaluations are conducted on the simulated datasets.

\begin{table}[]
     \centering
        \resizebox{\columnwidth}{!}{
  \setlength{\tabcolsep}{4mm}
  	\renewcommand{\arraystretch}{0.8}
    \begin{tabular}{ccccc}
    \midrule[0.8pt]
     \rowcolor[rgb]{ .949,  .949,  .949}  & \multicolumn{1}{c}{Baseline } & \multicolumn{1}{c}{Concat} & \multicolumn{1}{c}{ISFF} & \multicolumn{1}{c}{FFT-SF (Ours)} \bigstrut\\
       \midrule[0.8pt]
        Params &      1.33M    &    1.47M   &    1.35M   &  1.35M \bigstrut\\
    \hline
    GFLOPs &    22.41
  &   26.71  &   23.00    &  22.91 \bigstrut\\

    \hline
    PSNR  &    36.41  &   36.60    &  36.53     &  \textbf{36.95} \bigstrut\\
    \hline
    SSIM  &   0.954  &  0.959     &   0.959      &  \textbf{0.962}\bigstrut\\
    % \hline
    % SAM   &       &       &       &  \bigstrut\\
    \midrule[0.8pt]

    \end{tabular}%
    }
    \vspace{-4.5mm}
\caption{Ablation study of different stage fusion.}
  \label{tab:stg_ablation}%   
  \vspace{-2.5mm}
\end{table}
\begin{table}[]
\centering
 \resizebox{\linewidth}{!}{
  \setlength{\tabcolsep}{4mm}
  	\renewcommand{\arraystretch}{0.8}
   \vspace{-5mm}
    \begin{tabular}{ccccc}
    \midrule[0.8pt]
      \rowcolor[rgb]{ .949,  .949,  .949}   & Params & GFLOPs & PSNR  & SSIM \bigstrut\\
    \midrule[0.8pt]
      Sharing    &     0.46M       & 22.91      &  {35.65}           & {0.953} \bigstrut\\
    \hline
       No-Sharing   &   1.35M    &    22.91   &           \textbf{36.95}         &  \textbf{0.962}\bigstrut\\
   \midrule[0.8pt]
    \end{tabular}%
    }
    \vspace{-3mm}
      \caption{Ablation study of the independent network parameters and parameter sharing.}
      \vspace{-1mm}
\label{tab:sharing}
\end{table}

\noindent \textbf{Break-down Ablation.}
Here, we present the break-down ablation experiments on each component of our proposed framework. The Baseline is derived by removing the FFT-SF and pixel adaptive (PA) estimation module.  Denoisier is replaced by Restormer~\cite{zamir2022restormer}. The experimental results are shown in 
Table \ref{tab:bre_ablation}. It shows that the use of pixel-(PA) module leads to improved performance of 0.18 dB (from 36.19 dB to 36.37 dB). When we take the FFT-SF out, there is a drop of PSNR with 0.37 dB. Removing all the components, the performance degrades from 36.95 dB to 36.19 dB.
% \begin{figure*}[t]
% \scriptsize 
% \centering	
% \includegraphics[width=1\linewidth]{imgs/temp.png}
% \caption{\textcolor{red}{Reconstructed simulation HSI comparisons.}}
% \vspace{-2mm}
% \end{figure*}

% \begin{figure*}
%  \centering
%  	\scriptsize
% \includegraphics[width=1\linewidth]{imgs/temp2.png}

% 	\caption{Visual quality comparison of real noisy HSI experiments on Urban dataset with bands 1, 108, 208.}
%     \vspace{-3mm}
% 	\label{fig:urban_dataset} 
% \end{figure*}

% \begin{figure*}
%  \centering
%  	\scriptsize

% 	\caption{Visual quality comparison of real noisy HSI experiments on Urban dataset with bands 1, 108, 208.}
%     \vspace{-3mm}
% 	\label{fig:urban_dataset} 
% \end{figure*}

\begin{figure}

	\centering
	\scriptsize
	\setlength{\tabcolsep}{0.02cm}
	\begin{tabular}{ccccc}
		
		\includegraphics[width=1.5cm]{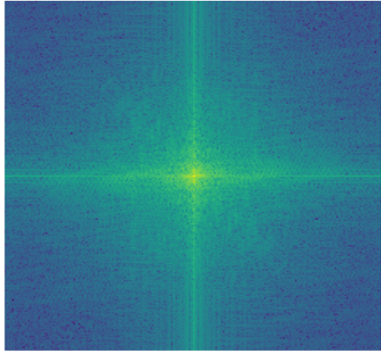}
		& \includegraphics[width=1.53cm]{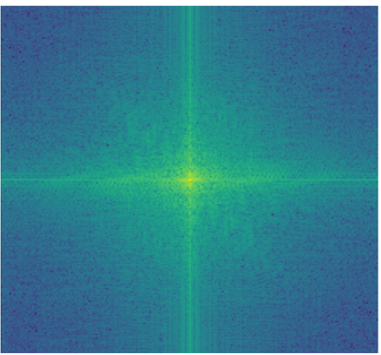}
		
		& \includegraphics[width=1.53cm]{imgs/fft/isff_fft.png}
		& \includegraphics[width=1.53cm]{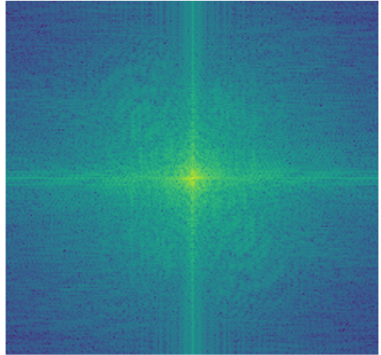}
		&		\includegraphics[width=1.51cm]{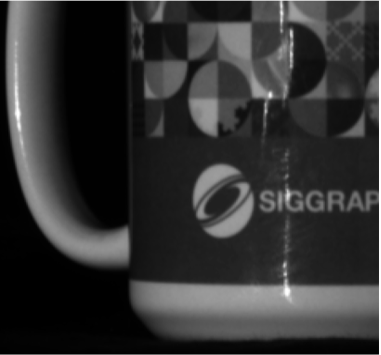}

		\\
		Concat & ISFF  & FFT-SF  & GT & Image \\

	\end{tabular}
	\vspace{-2mm}
	\caption{Visual comparison in the frequency domain when employing different fusion strategies.}
	\vspace{-2mm}
	\label{fig:frequency} 
\end{figure}

\noindent \textbf{Ablation Study of Stage Interaction.}
In Table. \ref{tab:stg_ablation}, we evaluate the efficacy of the proposed FFT-SF. We compare FFT-SF with two other strategies of fusion, \ie, directly Concat and ISFF~\cite{mou2022deep}. One could see that the stage interaction is important to deep unfolding framework, since it prevents critical information loss. And our proposed FFT-SF obtains the best results. Since FFT-SF conducts the fusion in the frequency domain, it can restore better high-frequency details as shown in Figure \ref{fig:frequency}. 

\noindent \textbf{Parameters Sharing.} To further demonstrate our method, we conduct the ablation study on the parameter-sharing network. Specifically, the modules in different stages shares the same parameters. The results are shown in Table \ref{tab:sharing}. Since our method employs the stage-targeted recovery by the pixel-adaptive data module, performance degrades when the weights of data module and denoisiers are shared.

\section{Conclusion}
In this paper, we propose a pixel adaptive deep unfolding transformer for HSI reconstruction. Our method aims to tackle the issues in existing deep unfolding works. In the data module, we employ the pixel-adaptive recovery, focusing on the imbalanced and agnostic degradation in CASSI. In the prior module, we introduce the Non-local Spectral Transformer to restore the HSI to emphasize the 3D characteristics. Moreover, inspired by the diverse expression of features in different stages and depths, the stage interaction is improved by the interaction in frequency domain. Experimental results reveal that our method surpasses the performance of the state-of-the-art HSI reconstruction methods. 

 {\small 
\noindent\textbf{Acknowledgments} This work was supported by the National Natural Science Foundation of China (62171038, 61931008, 62006023, and 62088101), and the Fundamental Research Funds for the Central Universities.
}

{\small
\bibliographystyle{ieee_fullname}
\bibliography{egbib}
}

\end{document}